\documentclass[onecolumn,11pt]{article}
\usepackage{pgf}
\usepackage{amsmath}
\usepackage{amssymb}
\usepackage{subfigure}
\usepackage{hyperref}
\usepackage{pgf}
\usepackage{bbm}
\usepackage{cite}
\usepackage{bbold}

\title{Learning an Optimization Algorithm through Human Design Iterations}
\usepackage{authblk}

\author[1]{Thurston Sexton\thanks{tbsexton@asu.edu}}
\author[1]{Max Yi Ren\thanks{yiren@asu.edu}}
\affil[1]{Department of Mechanical Engineering, Arizona State University}

\iftrue 
\newcommand{\todo}[1]{{\textcolor{red}{[[TODO: {#1}]]}}}
\newcommand{\commenttext}[1]{{\textcolor{red}{[[{#1}]]}}}
\newcommand{\highlight}[1]{{{{#1}}}}
\newcommand{\highlightrev}[1]{{{{#1}}}}
\newcommand{\change}[1]{{{{#1}}}}
\newcommand{\commentfoot}[1]{\footnote{\textcolor{red}{\emph{Comment: #1}}}}
\newcommand{\topic}[1]{}
\else
\newcommand{\todo}[1]{}
\newcommand{\commenttext}[1]{}
\newcommand{\commentfoot}[1]{}
\newcommand{\topic}[1]{}
\fi

\iffalse
        \newcommand{\cutsectionup}{\vspace*{-0.1in}}

        \newcommand{\cutsubsectionup}{\vspace*{-0.09in}}

        \newcommand{\cutsubsubsectionup}{\vspace*{-0.17in}}

        \newcommand{\cutparagraphup}{\vspace*{-0.1in}}

        \newcommand{\cutequationup}{\vspace*{-0.12in}}
        \newcommand{\cutequationdown}{\vspace*{-0.12in}}

\else 
        \newcommand{\cutsectionup}{}

        \newcommand{\cutsubsectionup}{}

        \newcommand{\cutsubsubsectionup}{}

        \newcommand{\cutparagraphup}{}

        \newcommand{\cutequationup}{}
        \newcommand{\cutequationdown}{}

\fi

\graphicspath{ {figures/} }

\begin{document}
\maketitle
\begin{abstract}
\change{Solving optimal design problems through crowdsourcing faces a dilemma: On one hand, human beings have been shown to be more effective than algorithms at searching for good solutions of certain real-world problems with high-dimensional or discrete solution spaces; on the other hand, the cost of setting up crowdsourcing environments, the uncertainty in the crowd's domain-specific competence, and the lack of commitment of the crowd, all contribute to the lack of real-world application of design crowdsourcing. We are thus motivated to investigate a solution-searching mechanism where \highlight{an optimization algorithm is tuned based on human demonstrations on solution searching}, so that the search can be continued after human participants abandon the problem. To do so, we model the iterative search process as a Bayesian Optimization (BO) algorithm, and propose an inverse BO (IBO) algorithm to find the maximum likelihood estimators of the BO parameters based on human solutions. We show through a vehicle design and control problem that the search performance of BO can be improved by recovering its parameters based on an effective human search. Thus, IBO has the potential to improve the success rate of design crowdsourcing activities, by requiring only good search strategies instead of good solutions from the crowd. }
\end{abstract}

\cutsectionup
\section{Introduction}
\label{sec:intro}
\change{\subsection{Challenges and opportunities for design crowdsourcing}}
\change{Optimal design problems often have large solution spaces and highly non-convex objectives and constraints, inhibiting effective solution searching through existing optimization algorithms. Some of these problems, however, have been quite successfully (yet heuristically) solved by human beings. Notable examples include protein folding~\cite{foldit,khatib2011algorithm}, RNA synthesis~\cite{eterna,lee2014rna}, genome sequence alignment~\cite{kawrykow2012phylo}, robot trajectory planning~\cite{sung2015robobarista}, and others~\cite{le2013crowdsourcing,ren2016ecoracer,schrope2013solving}. The superior performance of {\it some} human beings at solving these problems demonstrates the advantages of human intelligence, which are supported by cognitive science and neuroscience findings~\cite{lake2016building} (see discussion in Sec.~\ref{sec:limit}). However, despite a handful of success stories, applications of crowdsourcing to real-world design problems have yet to overcome several practical barriers. The cost of setting up problem-dependent crowdsourcing environments, the lack of commitment from crowd members, and uncertainty in domain-specific crowd competence have all contributed to its lack of adoption, while the growing availability of computation resources often makes straight-forward optimization or brute-force search a more convenient approach.} 

\begin{figure}
    \centering
    \includegraphics[width=1.0\linewidth]{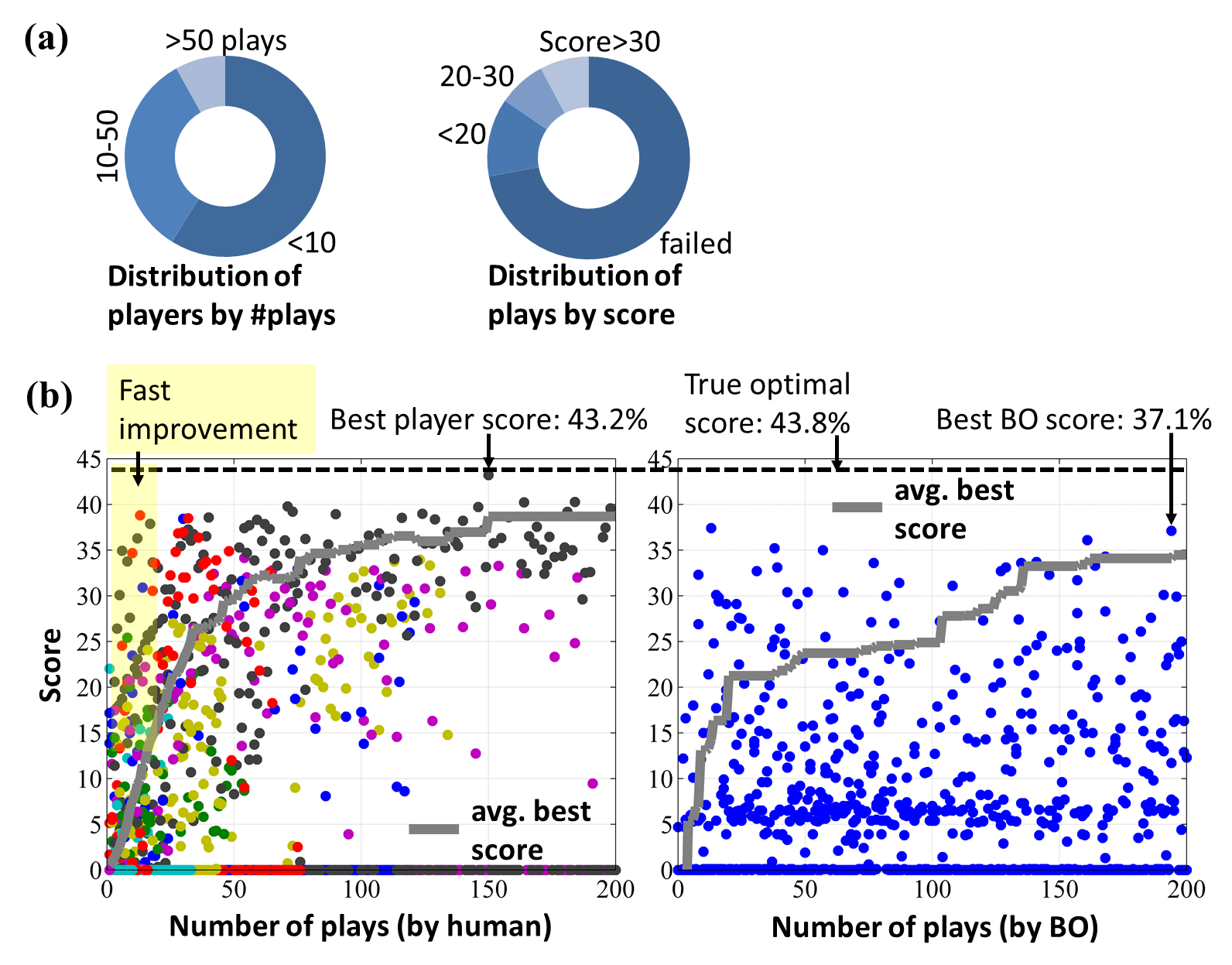}
    \caption{(a) Summary of player participation and performance (b) Results from the game showed while most players failed to outperform the Bayesian Optimization algorithm, some of them can identify good solutions early on. Image is reproduced from \cite{ren2015ecoracer,ren2016ecoracer}.}
    \label{fig:ecoracer}
\end{figure}

\change{Our earlier study~\cite{ren2016ecoracer} highlighted these challenges for design crowdsourcing: We gamified a vehicle design and control problem (called the ``ecoRacer'' problem \highlight{in what follows}) where the objective is to complete a track with the minimal energy consumption within a time limit, by finding the optimal final drive ratio of the vehicle and the control policy for acceleration and regenerative braking. The game was broadcast on social media and received more than 2000 plays from 124 unique players within the first month. Results showed that (1) the marginal improvement in average game score of the crowd over an algorithm does not necessarily justify the high cost for developing crowdsourcing games, and (2) only a few players were committed to the search for more than 50 iterations, and still fewer can outperform the computer-found solution at all (see summary in Fig.~\ref{fig:ecoracer}).} 

Nonetheless, human search results displayed a significantly different \textit{search pattern} than that of the algorithm. In particular, quite a few players showed rapid early improvement in performance, beyond the average performance of the computer, before they quit the game without reaching a solution close to the theoretical optimum. This observation is consistent with existing research (see, for example, \cite{khatib2011algorithm} on a human-designed protein folding algorithm having a short-term advantage over a standard algorithm), and suggests that while few people care to actually find the ``best solution", their early demonstrations on \textit{how they search} for a better solution may still be valuable. 
\highlight{Specifically, we hypothesize that if a computer algorithm can be tuned to mimic these demonstrations, it can serve as a replacement to human solvers in their absence, to search in an effective way without ever abandoning the problem.}  


\cutsubsectionup
\subsection{Learning to search}
\highlight{This paper aims to test the above hypothesis}. 
We model a human solver's search \highlight{behavior through} a Bayesian Optimization algorithm (BO, also known as Efficient Global Optimization)~\cite{jones1998efficient,brochu2010tutorial}. The algorithm iterates between two steps: (1) Estimating the shape of the problem space, based on previous solutions and corresponding performances, using a Gaussian Process (GP) model~\cite{rasmussen2006gaussian}, and (2) creating a new solution based on this estimate (details in Sec.~\ref{sec:pre}).
While BO is not provably the underlying mechanism humans use, we hypothesize that the algorithm can be tuned to mimic the results of successful human search strategies, specifically in comparison with other popular gradient- and non-gradient-based optimization algorithms. The key assumption in modeling human search behavior through BO is the use of a GP to account for human beings' learning of input-output relationships (or called ``function learning'' in psychology). This assumption is supported by various findings: In a recent review of function-learning models, Christopher et al.~\cite{lucas2015rational} showed that the two major schools of models, i.e., rules- and similarity-based, can be unified through a Gaussian Process\footnote{To be more accurate, the discussion in \cite{lucas2015rational} is for function learning with continuous variables. While our case study involves discrete variables (acceleration and braking signals), the dimension reduction process converts these variables to continuous ones. See Sec.~\ref{sec:real}.}. 
As discussed in Wilson et al.~\cite{wilson2015human}, the evidence that Occam's Razor plays an important role in human prediction also suggests that GP is an appropriate model for function learning, as GP reduces model complexity by construction~\cite{rasmussen2001occam}. Empirically, Borji et al.~\cite{borji2013bayesian} showed that BO, with the use of GP, has the closest convergence performance to human searches when applied to 1D optimization problems. In fact, many higher-dimensional problems that human beings naturally solve, such as locomotion planning, have also been successfully solved through the use of GP~\cite{levine2011nonlinear,deisenroth2013survey,calandra2014bayesian,cully2015robots}.

Under this modeling assumption, we investigate how BO parameters can be estimated for the algorithm to best match human solver's search trajectory, i.e., the sequence of solution-performance pairs. To this end, we introduce an Inverse BO (IBO) algorithm to derive the maximum likelihood estimators for BO parameters, and discuss challenges in its implementation (see Sec.~\ref{sec:est}). Validation of the IBO algorithm takes two steps. We first use a simulation study to show that IBO can successfully estimate BO parameters used in generating a search trajectory (Sec.~\ref{sec:sim}). We then show through the ecoRacer problem that the search performance of BO can be improved when its parameters are modified based on observing an effective human search and implementing IBO (Sec.~\ref{sec:real}). The results provide evidence that IBO can accelerate a search using only good search \textit{strategies} without needing a large number of good human solutions. Thus, incorporating IBO in design crowdsourcing may lower the requirement on crowd commitment and so increase its chance of success.  Limitations and their potential relaxations of the current IBO implementation will be discussed in depth in Sec.~\ref{sec:disc}.

\cutsubsectionup
\highlight{\subsection{Related work}
It is important to note that the focus of this paper is on the design of optimization algorithms aided by human demonstrations, rather than the derivation of qualitative explanations of the strengths and limitations of human design strategies. There have been numerous studies from the latter category in recent years (see \cite{pretz2008intuition,linsey2010study,daly2012design,cagan2013empirical,bjorklund2013initial,egan2016human} for example). This paper is also distinguished from studies that propose human-inspired optimization algorithms (see \cite{cagan1997simulated,landry2011protocol,mccomb2016drawing} for example), in that the learning of the optimization algorithm in our case is conducted by another algorithm, rather than by human researchers. From this aspect, our study is related to studies in learning-to-learn~\cite{thrun1998learning} where algorithms (e.g., for gradient-based optimization~\cite{wang2016learning} and optimal control~\cite{andrychowicz2016learning}) are tuned and controlled by a higher-level algorithm. In such work, however, the algorithms are often improved purely computationally through reinforcement learning by solving similar problems repeatedly. Due to the use of human demonstrations, our paper is also related to inverse reinforcement learning (see discussion in Subsec.~\ref{sec:mdp}), where human control strategies are used for defining and finding optimal control strategies.}

\cutsectionup
\section{Preliminaries on Bayesian Optimization}
\label{sec:pre}
\change{This section provides some background knowledge on BO to facilitate the discussion on IBO in Sec.~\ref{sec:est}.} 
\cutsubsectionup
\subsection{Terminologies and notations}
Let an optimization problem be $\min_{{\bf x}\in \mathcal{X}} f({\bf x})$ where $\mathcal{X} \subseteq \mathbb{R}^p$ is the {\it solution space}.
A search trajectory with $K$ iterations can be represented by $h_K := <{\bf X}_K, {\bf f}_K>$, where ${\bf X}_K$ and ${\bf f}_K$ represent the collection of $K$ samples in $\mathcal{X}$ and their objective values\highlight{, respectively}. $h_0 := <{\bf X}_0, {\bf f}_0>$ represents an initial exploration set with $K_0$ samples. Human strategy is represented by algorithmic parameters $\boldsymbol{\lambda}$ that govern the search behavior: During the search, each new solution ${\bf x}_{k+1}$ (for $k=0,\cdots,K-1$) is determined by $h_k:=<{\bf X}_k, {\bf f}_k>$ and $\boldsymbol{\lambda}$ through maximizing a merit function with respect to ${\bf x}$: ${\bf x}_{k+1} = \text{argmax}_{{\bf x}\in \mathcal{X}}Q({\bf x}; h_k, \boldsymbol{\lambda})$. The functional forms of the merit function $Q({\bf x})$ will be introduced in Subsecs.~\ref{sec:bo} and \ref{sec:est}. We also define $\boldsymbol{\Lambda}:=\text{diag}(\boldsymbol{\lambda})$ and \highlight{its estimator as $\boldsymbol{\hat{\Lambda}}:=\text{diag}(\boldsymbol{\hat{\lambda}})$}. 
\cutsubsubsectionup
\subsection{The BO algorithm}
\label{sec:bo}
\highlight{We briefly review the BO algorithm, to explain how each new sample ${\bf x}$ is drawn based on the merit function $Q({\bf x})$, itself defined by previous samples. Knowing this procedure is necessary for understanding the inverse BO algorithm, where we estimate the most likely BO parameters for a given trajectory of samples.} 

\begin{figure}
    \centering
    \includegraphics[width=0.9\linewidth]{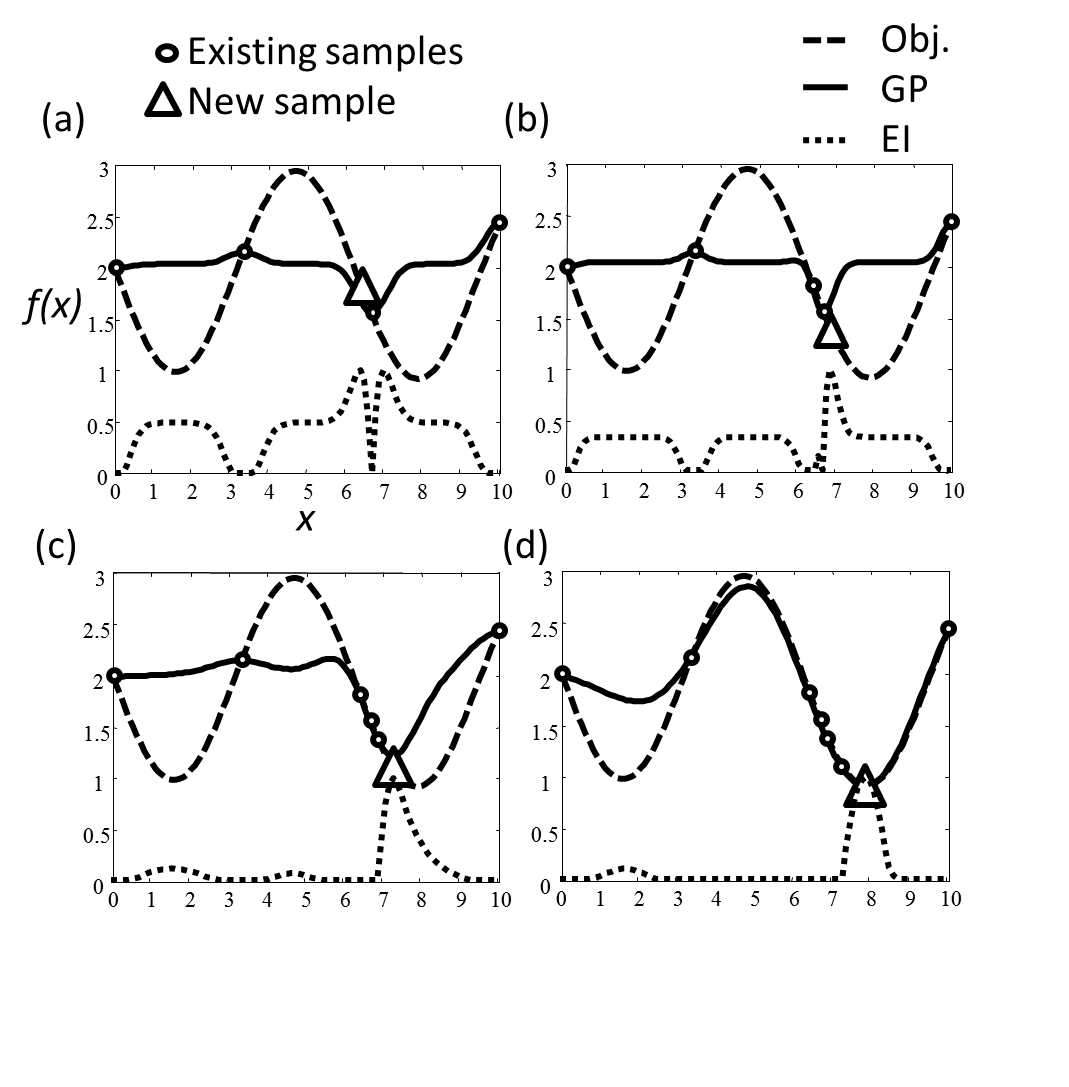}
    \caption{Four iterations of BO on a 1D function. Obj: The objective function. GP: Gaussian Process model. EI: Expected Improvement function. Image is modified from~\cite{ren2015ecoracer}.}
    \label{fig:bo}
\end{figure}

\highlight{BO contains two major steps in each iteration: For a collection of samples of a black-box function, a Gaussian Process (GP) model is updated; the merit function is then formulated based on the GP model, and the next sample is chosen by maximizing the merit.} {\bf Model update}: It first updates a Gaussian Process (GP) model to predict objective values, based on current observations $h_k$ and Gaussian parameters $\boldsymbol{\lambda}$. Without considering random noise in evaluating the objective, the GP model can be derived as $\hat{f}({\bf x};h_k,\boldsymbol{\lambda}) = b + {\bf r}^T{\bf R}^{-1}({\bf f}_k-b)$, where $b = \frac{{\bf 1}^T{\bf R}^{-1}{\bf f}_k}{{\bf 1}^T{\bf R}^{-1}{\bf 1}}$, ${\bf r}$ is a column vector with elements $r_i = \exp\left(-({\bf x}-{\bf x}_i)^T\boldsymbol{\Lambda}({\bf x}-{\bf x}_i)\right)$ for $i=1,\cdots,k$, ${\bf R}$ is a symmetric matrix with $R_{ij} = \exp\left(-({\bf x}_i-{\bf x}_j)^T\boldsymbol{\Lambda}({\bf x}_i-{\bf x}_j)\right)$ for $i,j=1,\cdots,k$, and ${\bf 1}$ is a column vector with ones. Without prior knowledge, the Maximum Likelihood Estimator (MLE) of $\boldsymbol{\lambda}$ for the GP model can be derived by solving
\cutequationup
\begin{equation}
    \boldsymbol{\hat{\lambda}}_{GP} = \text{argmin}_{\boldsymbol{\lambda}} \log (\sigma^k|{\bf R}|^{\frac{1}{2}}),
    \label{eq:mle}
\cutequationdown
\end{equation}
where $\sigma^2=({\bf f}_k-{\bf 1}b)^T{\bf R}^{-1}({\bf f}_k-{\bf 1}b)/n$ is the MLE of the GP variance. {\bf Sampling the solution space}: The second step is to determine the next sample using the GP model. A common sampling strategy is to pick the new solution in $\mathcal{X}$ that maximizes the {\it expected improvement} from the current best objective value $f_{\text{min}}:=\min {\bf f}_k$ (assuming a minimization problem): $Q_{EI}({\bf x};h_k,\boldsymbol{\lambda}) = (f_{\text{min}}-\hat{f})\Phi\left(\frac{f_{\text{min}}-\hat{f}}{\sigma}\right) + \sigma\phi\left(\frac{f_{\text{min}}-\hat{f}}{\sigma}\right)$. Here $\Phi(\cdot)$ and $\phi(\cdot)$ are the cumulative distribution function and probability density function of the standard normal distribution, respectively. The new sample is thus obtained by solving 
\cutequationup
\begin{equation}
    {\bf x}_{k+1} = \text{argmax}_{{\bf x}\in \mathcal{X}}Q_{EI}({\bf x};h_k,\boldsymbol{\lambda}).
    \label{eq:EI}
\cutequationdown
\end{equation}
Fig.~\ref{fig:bo} demonstrates four iterations of BO in optimizing a 1D function, with the GP model and the expected improvement function updated in each iteration. Note that similar to human searching behavior, BO is a stochastic process: First, the choice of the new design is stochastic, with better designs being more probable to be chosen\footnote{Numerically, this is because optimizing the non-convex function $Q_{EI}$ requires a nested global optimization routine, such as Genetic Algorithm (GA), CMA-ES~\cite{hansen2003reducing}, DIRECT~\cite{jones1993lipschitzian}, and BARON~\cite{sahinidis1996baron}. Some implementations of these, e.g., GA and CMA-ES, can be stochastic.}; and secondly, the initial exploration $h_0$ can be stochastic when it is modeled by a random sampling scheme, e.g., Latin Hypercube sampling (LHS, see \cite{jones1998efficient} for details).

\cutsectionup
\section{Inverse BO}
\label{sec:est}

\change{We consider human solution search to consist two stages: A few exploratory searches are first conducted to acquire a preliminary understanding of the problem, before the execution of BO follows. For example, a player may spend a few trials to get familiar with a new game, before thinking about strategies to improve his score. IBO minimizes the sum of two costs corresponding to the exploration and BO stages, respectively. By doing so, it finds the most likely explanation of the underlying search strategy.}

Specifically, IBO estimates $\boldsymbol{\lambda}$, along with the size of the initial exploration set $K_0$, given the trajectory $h_K$. To do so, we introduce and minimize a cost function consisting of the exploration cost for $h_0$, denoted as $L_{INI}$, and the BO cost for the rest of $h_K$, denoted as $L_{BO}$. We define $L_{INI}:=-\log\left(Dp({\bf X}_0)\right)$ where $p({\bf X}_0)$ is the joint probability of the exploration set and $D:=|\mathcal{X}|$ is the size of the solution space; and $L_{BO}:=-\log\left(Dp(h_K-h_0|h_0)\right)=-\sum_{k=0}^{K-1} \log\left(Dp({\bf x}_{k+1}|h_k)\right)$ where $p({\bf x}_{k+1}|h_k)$ is the density for choosing ${\bf x}_{k+1}$ conditioned on $h_k$. Here $\log(\cdot)$ stands for natural logarithm. 

The derivation of $L_{INI}$ and $L_{BO}$ are as follows: To calculate $L_{INI}$, we assume that each new sample during the exploration phase, ${\bf x}_i$ for $i=1,\cdots,K_0$, tends to maximize its minimum Euclidean distance $d({\bf x}_i,{\bf X}_{<i})$ to previous samples ${\bf X}_{<i}$, this is referred to as the {\it max-min sampling scheme} \highlight{in what follows}. Let the joint probability of the exploration set be $p({\bf X}_0) = p({\bf x}_1)p({\bf x}_2|{\bf x}_1)\cdots p({\bf x}_{K_0}|{\bf X}_{<K_0})$ and each conditional probability follow a Boltzmann distribution: $p({\bf x}_i|{\bf X}_{<i}) = \exp\left(\alpha_{INI} d({\bf x}_i,{\bf X}_{<i})\right)/Z_{INI}({\bf x}_i,\alpha_{INI})$. Here the scalar $\alpha_{INI}$ represents how strictly each sample from ${\bf X}_0$ follows the max-min sampling scheme, and $Z_{INI}({\bf x}_i,\alpha_{INI}) = \int_{{\bf x}\in\mathcal{X}}\exp\left(\alpha_{INI} d({\bf x},{\bf X}_{<i})\right)d{\bf x}$ is a partition function 
that ensures that $\int_{\mathcal{X}} p({\bf x}_i|{\bf X}_{<i})d{\bf x} = 1$. Note that the first sample in the exploration set is considered to be uniformly drawn, and thus its contribution to the cost (a constant) can be omitted. 

To calculate $L_{BO}$, the conditional probability density of sampling ${\bf x}\in \mathcal{X}$ based on current $h_k$ can be similarly modeled as a Boltzmann distribution:
\cutequationup
\begin{equation}
    p({\bf x}|h_k) = \exp\left(\alpha_{BO} Q_{EI}({\bf x};h_k,\boldsymbol{\lambda})\right)/Z_{BO}(h_k,\boldsymbol{\lambda},\alpha_{BO}),
    \label{eq:density}
\cutequationdown
\end{equation}
where $Z_{BO}(h_k,\boldsymbol{\lambda},\alpha_{BO}) = \int_{{\bf x}\in\mathcal{X}}\exp\left(\alpha_{BO} Q_{EI}({\bf x};h_k,\boldsymbol{\lambda})\right)d{\bf x}$ is also a partition function. The parameter $\alpha_{BO}$ plays a similar role to $\alpha_{INI}$. For simplicity, we define $\tilde{l}_i:=-\log\left(Dp({\bf x}_i|{\bf X}_{<i})\right)$ and $l_k:= -\log\left(Dp({\bf x}_{k+1}|h_k)\right)$, so that $L_{INI} = \sum_{i=1}^{K_0} \tilde{l}_i$ and $L_{BO} = \sum_{k=0}^{K-1} l_k$.  A lower value of $\tilde{l}$ or $l$ represents higher probability density of the current sample to be drawn by max-min sampling or BO, respectively, and a zero indicates that the sample can be considered as uniformly drawn. 

IBO solves the following problem to derive $\boldsymbol{\hat{\lambda}}$.
\cutequationup
\begin{equation}
   \min_{\alpha_{INI}, \alpha_{BO},\boldsymbol{\lambda}, K_0} L:=L_{INI}+L_{BO}
   \label{eq:obj}
\cutequationdown
\end{equation}
Note that to find the optimal $K_0$ for any given $\alpha_{INI}$, $\alpha_{BO}$, and $\boldsymbol{\lambda}$, one can first calculate the optimal $\tilde{l}_i$ and $l_k$ for $i,k=2,\cdots,K$, with respect to $\alpha_{INI}$, $\alpha_{BO}$, and $\boldsymbol{\lambda}$, and then scan $K_0 = 2,\cdots,K$ to find the lowest value of $L_{INI}+L_{BO}$. The scan starts at $K_0=2$ because it is not meaningful to initialize BO with a single sample. 
\cutsubsectionup
\subsection{Numerical Integration for $Z_{BO}$}
\label{sec:par}
The calculation of each $l$ requires an approximation of the integral $Z_{BO}(h_k,\boldsymbol{\lambda},\alpha_{BO})$, where the integrand $Q_{EI}({\bf x};h_k,\boldsymbol{\lambda})$ is usually a highly non-convex function with respect to ${\bf x}$, with function values dropping significantly around local maxima. See Fig.\ref{fig:bo} for example. Thus we propose to approximate $Z_{BO}$ with importance sampling using a customized proposal density function that combines a uniform distribution with density $p({\bf x}) = 1/D$ and a multivariate normal distribution with density $q({\bf x}) = (\sqrt{2\pi}\sigma_{I}^p)^{-1}\exp(-||{\bf x}-\boldsymbol{\mu}||^2/2\sigma_{I}^2)$, where $\sigma_{I}$ and $\boldsymbol{\mu}$ are parameters of $q(\bf x)$. The uniform distribution is used to sample over $\mathcal{X}$, while the normal distribution helps to improve the approximation by capturing the potential peak at the current sample ${\bf x}_{k+1}$. Thus we set $\boldsymbol{\mu}:={\bf x}_{k+1}$. Let ${\bf x}_i^u \in \mathcal{U}$ for $i=1,...,I$ and ${\bf x}_j^n \in \mathcal{N}$ for $j=1,...,J$ be samples from \highlight{$p({\bf x})$ and $q({\bf x})$}, respectively. The approximation $\hat{Z}_{BO}$ can be calculated by 
\cutequationup
\begin{equation}
    \hat{Z}_{BO} := \sum_{\mathcal{U}}\frac{DQ_{EI}({\bf x}_i^u)}{I\left(1+Dq({\bf x}_i^u)\right)} + \sum_{\mathcal{N}}\frac{DQ_{EI}({\bf x}_j^n)}{J\left(1+Dq({\bf x}_j^n)\right)},
    \label{eq:importance}
\cutequationdown
\end{equation}
with arguments of $Q_{EI}$ omitted for simplicity. The derivation of Eq.~\eqref{eq:importance} is deferred to the appendix. Note that this approximation works under the assumption that $\int_{{\bf x} \in \mathcal{X}} q({\bf x})d{\bf x} \approx 1$, which is plausible as the normal distribution is designed to have a narrow spread to match the local peak at ${\bf x}_{k+1}$. In this paper, the shape of this normal distribution is set by $\sigma_{I}=0.01$ universally. While the setting of $\sigma_{I}$ affects the variance of the approximation of $Z_{BO}$, we found this setting to perform well in practice. For $Z_{INI}$, since the minimum Euclidean distance function in a high dimensional space with limited samples is a relatively smooth function, we use Monte Carlo sampling for its approximation.

\cutsubsectionup
\subsection{Simulation studies}
\label{sec:sim}
\change{As a validation step, we show that IBO can recover the parameters of a general BO given only an observed search trajectory. If IBO can determine the correct parameters (1) after a few number of iterations, (2) in a high-dimensional problem space, and (3) from a wide range of trajectory/parameter settings,
then it \highlight{could be used} to recover parameters for matching a BO algorithm to an observed human search.}

We use a simulation study to show that, for a given search trajectory, IBO can correctly identify the true $\boldsymbol{\lambda}$ provided the trajectory is sufficiently different from a random search. In addition, the simulation indicates that learning from already-efficient search behavior (i.e., estimating $\boldsymbol{\lambda}$ through IBO of an observed effective search trajectory) can lead to better BO convergence than the more common self-improvement methods (i.e., updating $\boldsymbol{\hat{\lambda}}$ by maximizing the likelihood of the observations according to the GP model).

\cutsubsubsectionup
\subsubsection{Simulation settings and results}
The simulation study is detailed as follows: We apply BO to a 30-dimensional Rosenbrock function constrained by $\mathcal{X}:= [-2,2]^{30}$. To initialize BO, we use LHS to draw $10$ samples from $\mathcal{X}$. BO terminates when the expected improvement for the next iteration is less than $10^{-3}$. At each iteration, the expected improvement is maximized using a multi-start gradient descent algorithm~\cite{zhu1994bfgs} with 100 LHS initial guesses. A set of BO parameters, $\boldsymbol{\Lambda}=0.01 {\bf I}, 0.1 {\bf I},1.0 {\bf I}$, and $10.0 {\bf I}$, are used to perform the search, where ${\bf I}$ is the identity matrix. For each of the four settings, $30$ independent trials are recorded. 

\begin{figure}
    \centering
    \includegraphics[width=\linewidth]{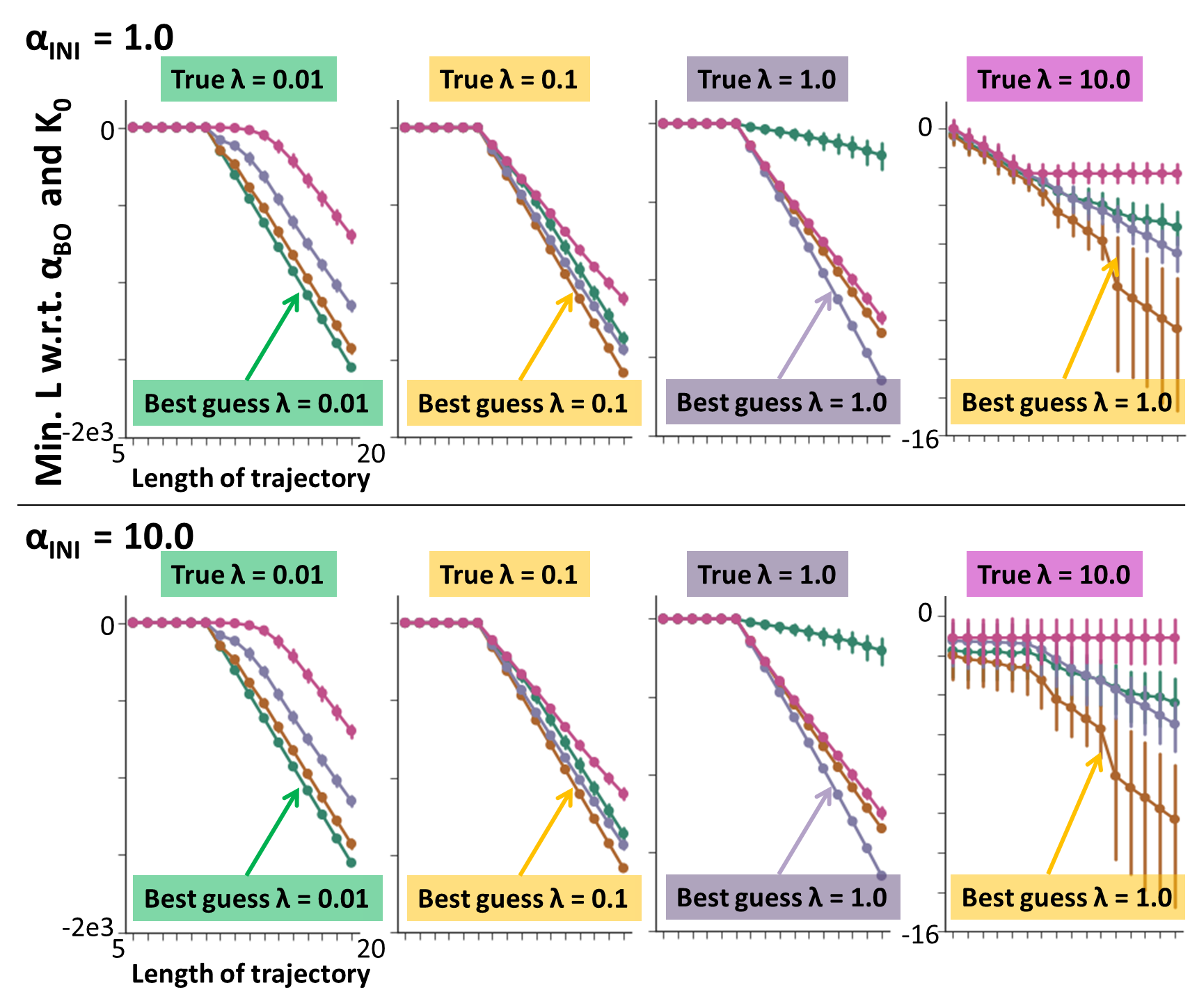}
    \caption{The minimal cost $L$ for search trajectory lengths $N = 5,...,20$ with respect to $\mathcal{G}_{\alpha_{BO}}$ and $\mathcal{G}_{K_0}$. $\alpha_{INI}$ is fixed to $1.0$ and $10.0$. \textit{View in color.}}
    \label{fig:simulation1}
\end{figure}

For each BO setting $\boldsymbol{\Lambda}$, each candidate estimator $\boldsymbol{\hat{\Lambda}}$, and each trajectory of length $K=5,...,20$, we solve Eq.~\eqref{eq:obj} using a grid search with $\mathcal{G}_{\alpha_{BO}} := \{0.01, 0.1,1.0,10.0\}$ and $\mathcal{G}_{K_0} := \{2,\cdots,K\}$. We fix $\alpha_{INI}$ to $1.0$ and $10.0$, and will discuss its influence to the estimation. 
Fig.~\ref{fig:simulation1} presents the resulting minimal $L$ for all four cases and under all guesses. Each curve in each subplot shows how the minimal $L$ (with respect to $\alpha_{BO}$ and $K_0$) changes as the search continues. The means and standard deviations of $L$ are calculated using the 30 trials. $Z_{INI}$ is approximated using a sample size of $10,000$. In approximating $Z_{BO}$, samples from the normal and the uniform distributions are of equal sizes ($I=J=5,000$).

\cutsubsubsectionup
\subsubsection{Analysis of the results}
Based on the results from this simulation, as summarized in Fig.~\ref{fig:simulation1}, the major finding from this simulation study is that IBO can successfully recover the BO parameters in cases where BO does not resemble uniform random sampling of the design space. In the cases of $\boldsymbol{\Lambda}=0.01 {\bf I}, 0.1 {\bf I},1.0 {\bf I}$, we see that the correct choices of $\boldsymbol{\hat{\Lambda}}$ consistently lead to the lowest cost along the search process. After only one or two iterations, in nearly all cases, the correct parameter has the highest likelihood of all four propositions, and this remains the case along the search.  However, under large BO parameters such as $\boldsymbol{\Lambda}=10.0 {\bf I}$, the similarity between any two points in the design space becomes close to zero, leading to (almost) uniform uncertainty and expected improvement. Therefore this setting reduces BO to a uniform random sampling scheme. Fig.~\ref{fig:simulation1}d shows that IBO does not perform well in this situation. To better understand the behavior of IBO under near-random searches, a curious reader may find a discussion on the properties of the costs $l$ and $\tilde{l}$ in the Appendix.

\cutsubsubsectionup
\subsubsection{Learning from others vs. self-adaptation}
\begin{figure}
    \centering
    \includegraphics[width=\linewidth]{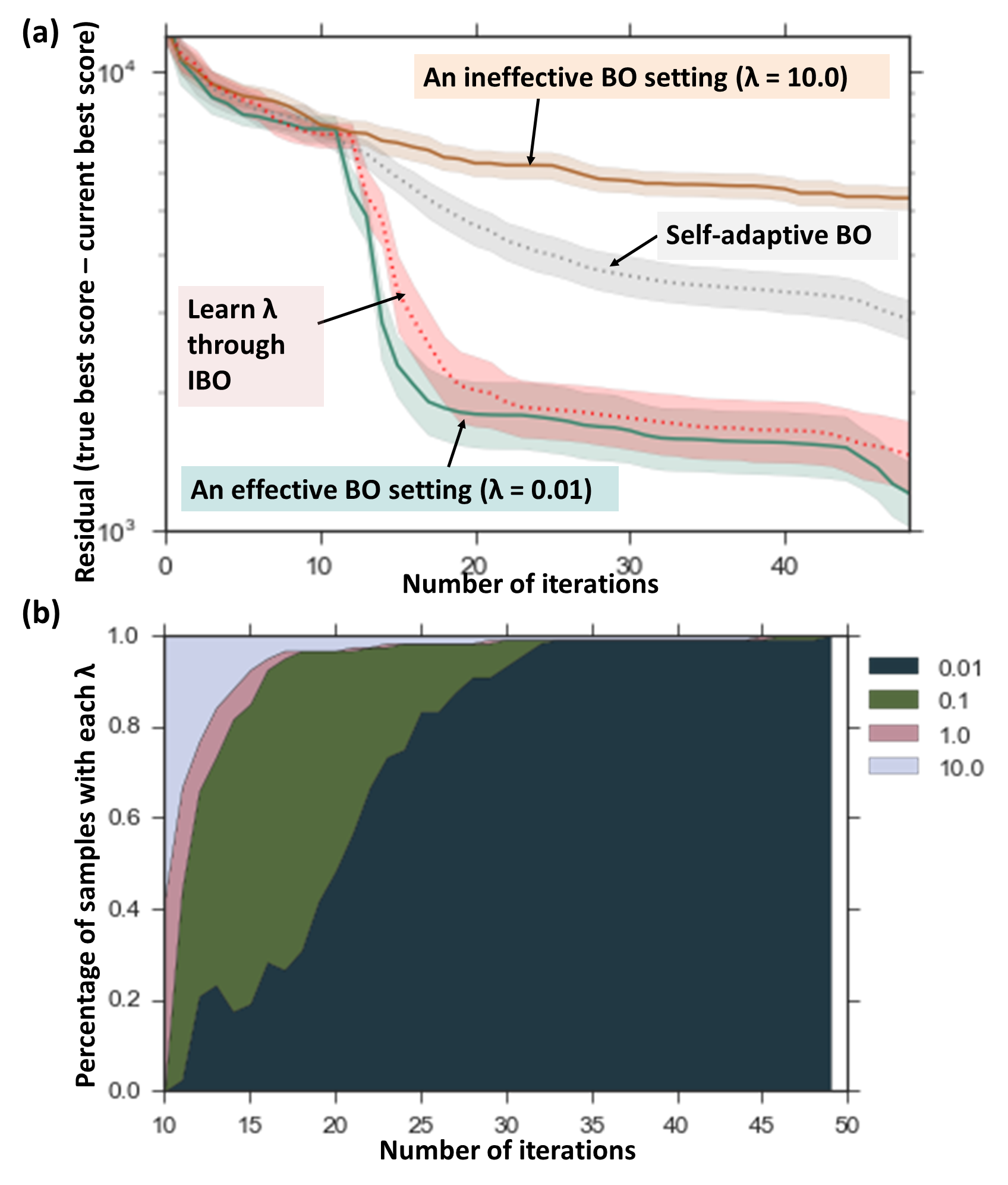}
    \caption{(a) Comparison on BO convergence using four algorithmic settings: (orange) $\boldsymbol{\Lambda} = 10.0{\bf I}$, (green) $\boldsymbol{\Lambda} = 0.01 {\bf I}$, (grey) the MLE of $\boldsymbol{\Lambda}$ is used for each new sample, and (red) the initial setting $\boldsymbol{\Lambda} = 10.0{\bf I}$ is updated by IBO using the trajectory from $\boldsymbol{\Lambda} = 0.01 {\bf I}$. (b) The percentages of estimated $\boldsymbol{\hat{\Lambda}_{MLE}}$ along the number of iterations, averaged over the cases with $\boldsymbol{\Lambda} = \{0.01 {\bf I}, 0.1 {\bf I},1.0 {\bf I},10.0 {\bf I}\}$ and 30 trials for each case. View in color.}
    \label{fig:simulation2}
\end{figure}

The above study showed that the correct BO setting $\boldsymbol{\lambda}$ can be learned through IBO. This subsection further demonstrates the advantage of ``learning from others'' (i.e., updating $\boldsymbol{\lambda}$ through IBO), over ``self-adaptation'' (i.e., finding the MLE of $\boldsymbol{\lambda}$ using $h_k$). The settings follow the above study and results are shown in Fig.~\ref{fig:simulation2}. First, to show the significant influence of $\boldsymbol{\lambda}$ on search effectiveness, we show the convergence of two fixed search strategies with $\boldsymbol{\Lambda}=0.01$ and $10.0$. Note that while neither converges to the optimal solution within 50 iteration, the former is significantly more effective than the latter. For ``self-adaptive BO'', we use a grid search ($\mathcal{G}_{\boldsymbol{\Lambda}} = \{0.01 {\bf I}, 0.1 {\bf I},1.0 {\bf I},10.0 {\bf I}\}$) to find $\boldsymbol{\hat{\Lambda}_{GP}}$ that maximizes Eq.~\eqref{eq:mle} at each iteration, and use $\boldsymbol{\hat{\Lambda}_{GP}}$ to find the next sample. We show in Fig.~\ref{fig:simulation2}b the percentages of the four guesses being $\boldsymbol{\hat{\Lambda}_{GP}}$ along the search, using $\mathcal{G}_{\boldsymbol{\Lambda}}$ as the initial guesses for BO. The ``learning from others'' case starts with $\boldsymbol{\Lambda}=10.0{\bf I}$ and uses IBO to derive $\boldsymbol{\hat{\Lambda}}$ from the trajectory produced by $\boldsymbol{\Lambda}=0.01{\bf I}$. From Figs.~\ref{fig:simulation1} and \ref{fig:simulation2}b, we see that $\boldsymbol{\hat{\Lambda}_{GP}}$ does not converge to $\boldsymbol{\Lambda} = 0.01{\bf I}$ as quickly as IBO, which explains why ``learning from others'' outperforms ``self-adaptation'' in Fig.\ref{fig:simulation2}a. It is worth noting that this difference in performance may be relatively dependant on the dimensionality of the problem, as the two strategies were found to have similar convergence performance when applied to 2D functions. One potential explanation for this is that, in a lower dimensional space, an effective $\boldsymbol{\hat{\Lambda}_{GP}}$ can be learned with a smaller number of samples. 

\cutsectionup
\section{Case study}
\label{sec:real}
We now investigate how IBO may improve the performance of BO when applied to a vehicle design and control problem. 

\cutsubsectionup
\subsection{Dimension reduction for player's control signals}
The solution data from each game play consists of (1) the final gear ratio, (2) the recorded acceleration and braking signals, and (3) the corresponding game score. The length of a raw control signal matches that of the track, which has 18160 distance steps. Encoding control signals to a low dimensional space is feasible since common acceleration and braking patterns exist across all plays. In \cite{ren2015ecoracer}, this was done by introducing manually defined state-dependent basis functions (i.e., polynomials of the velocity of the car, slope of the track, distance to the terminal, remaining battery energy, and time spent) to parameterize the control signals. The underlying assumption that human players are aware of all the state-dependent bases is untested.

\change{In this paper, we perform dimension reduction based on evidence that human beings often solve high dimensional problem by performing problem abstraction and using a hierarchical search~\cite{mcgovern1997roles,mcgovern2001automatic,dietterich1998maxq,kulkarni2016hierarchical,botvinick2014model}. In the context of the ecoRacer game, we hypothesize that players segment the track into $m$ discrete sections, and make separate control decisions in each segment.} 
Mathematically, this is equivalent to projecting observed signals onto $m$ independent basis,
which can be elegantly addressed by ICA~\cite{stone2004independent}. Compared with Principal Component Analysis, where the bases minimize the covariance of the data, our ICA implementation maximizes the Kullback-Leibler divergence between all bases pairs, and is more suitable for non-Gaussian signals, such as the control data from this game (i.e., the acceleration/braking signals across players at each step along the track are unlikely to follow a Gaussian distribution).
Much like PCA, the choice of the number of ICA bases requires a balance between fidelity and practicality. While it is theoretically possible to find the ``most likely" number of bases using information-theoretic criteria for model selection~\cite{hui2011empirical}\footnote{For completeness, we used 1000 PCA components as preprocessing to obtain the most likely number of ICA components under three suitable criteria: Minimum Description Length, Akaike Information Criterion, and Kullback Information Criterion, as 187, 464, and 373, respectively, using the method from \cite{hui2011empirical}. 
While these dimensionalities could make sense from a neurological perspective (e.g., given that the game takes 36 seconds, a decision interval of $36{\rm s}/187=192{\rm ms}$ is close to the range for the time-frame of attentional blink, which is 200-500 ms~\cite{tombu2011unified}), the resultant high-dimensional solution spaces are unfavorable for BO.}, we chose to use 30 bases because (1) over $95\%$ of the variance is explained, and (2) the resultant solution space (30 control variables and one design variable) is small enough for BO to be effective.



\begin{figure}
    \centering
    \includegraphics[width=0.9\linewidth]{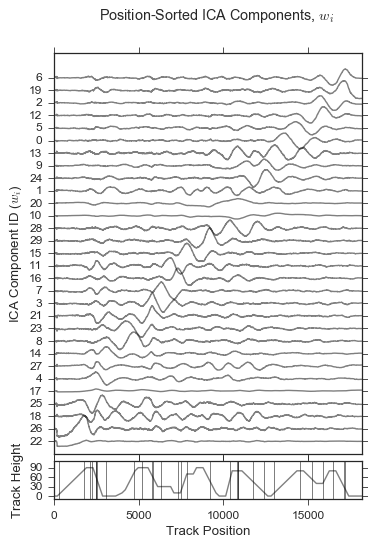}
    \caption{ICA bases learned from all human plays and the ecoRacer track. Vertical lines on the track correspond to the peak locations of the bases.}
    \label{fig:ica}
\end{figure}

\cutsubsubsectionup
\subsection{Derivation of $\boldsymbol{\hat{\lambda}}$ and $\boldsymbol{\hat{\lambda}}_{GP}$}
We apply IBO to two players, referred to as ``P2'' \highlight{and ``P3''}, who achieved the second \highlight{and third} highest score within 31 and 73 plays\highlight{, respectively}, much less than the 150 plays from the achiever of the highest score. To do so, we first encode all control solutions from \highlight{the two players} using the learned ICA bases. Together with the final drive ratios, all solutions are then normalized to be within $[-1,1]^{31}$. IBO is performed separately on P2 and P3. We found that the probability for either player to have followed the max-min sampling scheme is lower than that of following BO, as the minimal values of $\tilde{l}({\bf x}_k, \alpha_{INI})$ for $k=2,...,31$ (with respect to $\alpha_{INI}$) are dominated by those of $l({\bf x}_k, \alpha_{BO})$. This means that the \highlight{players were not likely to have performed an exploration before they started trying to improve their performance}. This finding is reasonable, as the scoring mechanism in ecoRacer game, just like in other racer games with fairly predictable vehicle dynamics, can be understood by the player early on. Therefore, the search for $\boldsymbol{\hat{\lambda}}$ is performed by solving Eq.~\eqref{eq:obj} with $\boldsymbol{\lambda} \in [0.01,10.0]^{31}$, $\alpha_{BO}\in \mathcal{G}_{\alpha_{BO}}$, and a minimal number of initial samples ($K_0 = 2$) required for BO. 
For comparison purpose, we obtain $\boldsymbol{\hat{\lambda}}_{GP}$ using plays from P2, \change{which represents a case where BO parameters are fine-tuned by the observed game plays, without trying to explain why these solutions were searched by the player.}    

Due to the non-convexity of Eq.~\eqref{eq:obj} and Eq.~\eqref{eq:mle}, gradient-based searches using a series of 10 initial guesses are conducted to avoid inferior local solutions. Finite difference is used for gradient approximation. Both $\boldsymbol{\hat{\lambda}}$ and $\boldsymbol{\hat{\lambda}}_{GP}$ are calculated offline, and fixed during the execution of BO. 

\cutsubsubsectionup
\subsection{Comparison of BO performance}
\begin{figure}
    \centering
    \includegraphics[width=\linewidth]{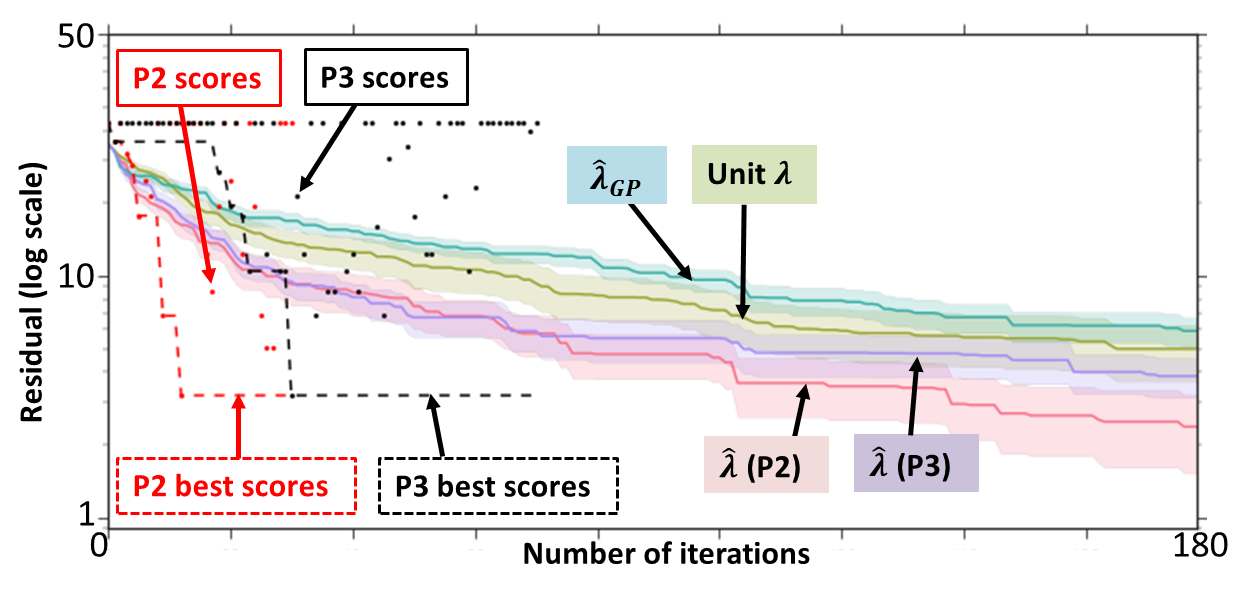}
    \caption{The residual of current best score vs. the known best score, with settings  $\boldsymbol{\hat{\lambda}}$ (IBO, red), $\boldsymbol{\hat{\lambda}}_{GP}$ (MLE, blue), and the default $\boldsymbol{\lambda} = {\bf I}$ (green). Results are shown as averages over 30 trials. One-sigma confidence intervals are calculated via $5000$ bootstrap samples. 
    Red and black dots are scores from P2 and P3, respectively. 
    }
    \label{fig:results}
\end{figure}

Fig.~\ref{fig:results} compares the BO performance under $\boldsymbol{\hat{\lambda}}$ (for P2 and P3), $\boldsymbol{\hat{\lambda}}_{GP}$ and $\boldsymbol{\Lambda}={\bf I}$. In each case, we start with the first two plays from the players, and run 180 BO iterations. Similar to the simulation study, results are reported using 20 trials due to the stochastic nature of BO. Due to the small trial number, bootstrap variance estimators are reported as the shades around the average in the figure. $\boldsymbol{\hat{\lambda}}$ outperforms the other two settings consistently along the search with statistical significance. The BO performance by mimicking P2 is slightly better than that of P3.

The result shows that BO can be improved noticeably by learning from P2 and P3. However, \highlight{the players' search are not fully mimicked by IBO, as they improved much faster than the modified BO does}, indicating that the proposed model still has room for improvement. Nevertheless, \highlight{the IBO implementation still achieves the closest performance to the players'} among all BO instances, and it is the only algorithm that achieved better performance than the players' best play within 100 iterations. This result demonstrates the potential of IBO to continue an effective human search after the player quits, with an improved search performance from a standard BO.

\begin{figure}
    \centering
    \includegraphics[width=\linewidth]{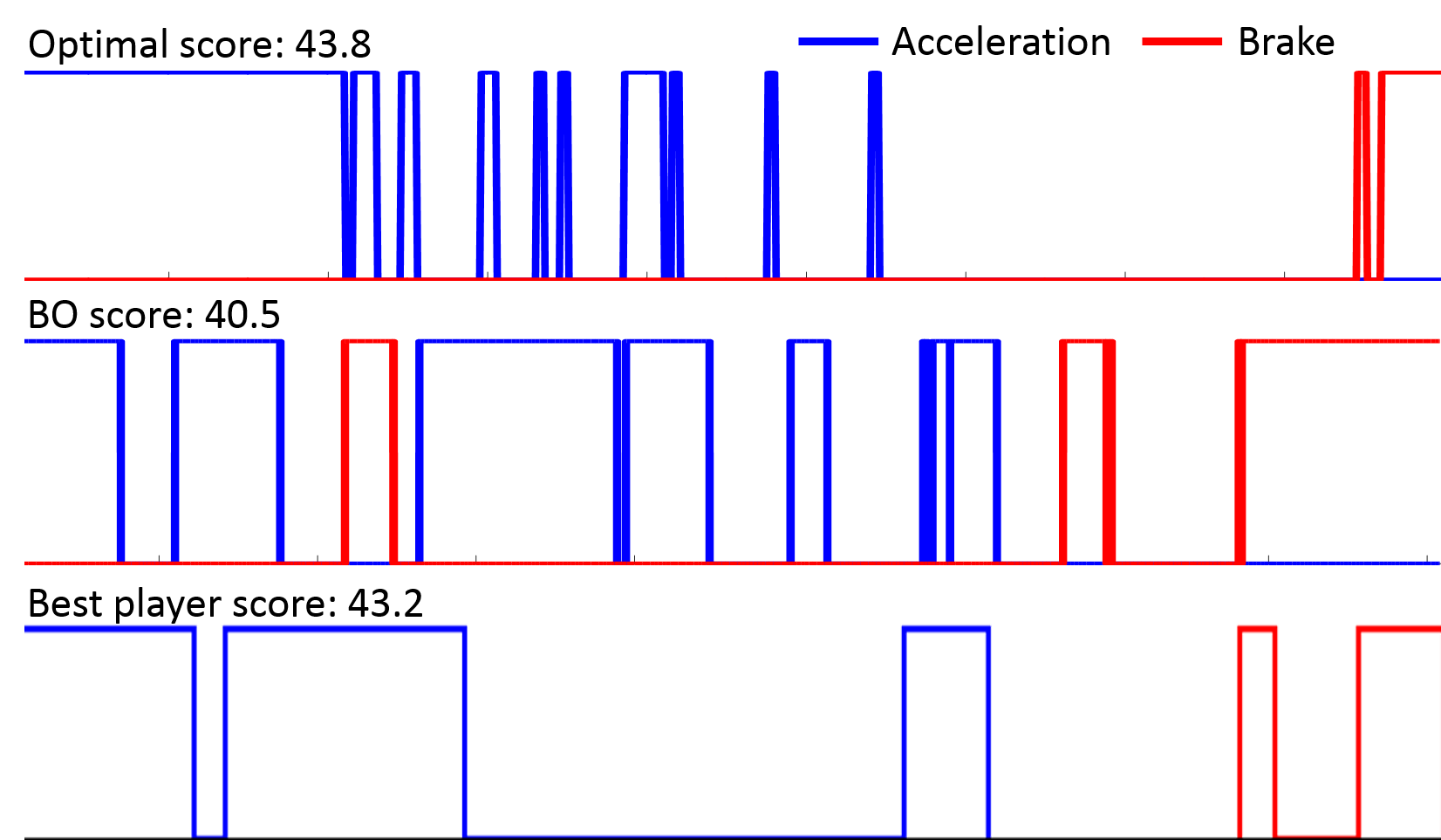}
    \caption{Qualitative comparison on control strategies from the theoretical optimal solution (top), one of the BO solutions (middle), and the best player solution (bottom).}
    \label{fig:compare_strategy}
\end{figure}

\highlightrev{For completeness, we also note that in all cases, the BO identifies the true optimal final drive ratio at the end of the search. We also qualitatively compare the best human solution with one BO solution with high score, along with the theoretically optimal solution in Fig.~\ref{fig:compare_strategy}. The result indicates that while these control strategies yield similar scores, they are quantitatively different, although braking towards the end is observed as a common strategy. Human search data are documented at \url{ecoracer.herokuapp.com/results}, where the best players' solution strategies are published.}

\cutsectionup
\section{Discussion}
\label{sec:disc}
The above study provided a starting point for learning optimization algorithms based on human solution-search data. Yet, many pressing questions remain unanswered. This section will address a few notable ones. \highlight{Some potential answers to these questions will rely on readers' familiarity with Inverse Reinforcement Learning~\cite{ng2000algorithms,ziebart2008maximum,levine2011nonlinear} (IRL, also called apprenticeship learning~\cite{abbeel2004apprenticeship,abbeel2010autonomous} and inverse optimal control~\cite{dvijotham2010inverse}). To familiarize readers with this topic, a discussion on the connection between IBO and IRL is provided in Subsec.~\ref{sec:mdp}.}

\cutsubsectionup
\subsection{Limitations and potential values of IBO}
\label{sec:limit}
From the case study, a strategy learning through IBO outperformed default algorithms, but is yet to reach the performance of the best human solver. This indicates potential room to further improve the algorithm. In the following, we discuss notable limitations of IBO. We shall also note that these also apply to the general problem of designing optimization algorithms through human demonstrations (called DO in what follows). 

{\bf Model of human search strategies}: Studies in cognitive science have put forth several core ingredients of human intelligence, including intuitive physics~\cite{spelke1995development,baillargeon2009account,bates2015humans,gershman2015computational}, problem decomposition skills~\cite{fodor1975language,biederman1987recognition,kulkarni2016hierarchical}, ability in learning-to-learn~\cite{harlow1949formation}, and others~\cite{lake2016building}. 
While evidence has shown the connection between BO and human search~\cite{borji2013bayesian}, suitable models for human search strategies can be problem dependent. For example, for low-dimensional design problems, Egan et al.~\cite{egan2015synergistic} showed that people adopting univariate search are more likely to achieve effective search. \highlight{This result is supported by earlier psychological studies on how children perform scientific reasoning, and thus may be useful to explain how people identify unfamiliar systems.} However, univariate search may not reflect how people search for solutions in a familiar context (such as car driving) and with a large number of control and design variables to tune, \highlight{as is the situation of the ecoRacer game}.
\highlight{For such high-dimensional and physics-based design and control problems}, a potentially reasonable human search model could be to incorporate human intuitive physics models into the evaluation of the expected improvement. Thus instead of estimating GP parameters, one could estimate a statistical model of the state-space equations of the dynamical system, which influences the expected improvement. At a more abstract level, the fundamental challenge in understanding how a human search strategy should be modeled is the lack of knowledge about the functional form of the local objective (i.e., the Q-function) that governs the generation of new solutions during the search based on the current state (cumulative knowledge learned by the human solver). \highlight{As we will discuss later in this section, this challenge is also a key topic in IRL. Not surprisingly, one notable solution from IRL to this problem is in fact to use non-parametric models such as GP~\cite{levine2011nonlinear,choi2012nonparametric}.}

\change{{\bf Uncertainty in estimation}: A limited amount of demonstrations could be insufficient to provide a good estimation of the BO parameters, even though the underlying parameters are the effective ones.
One potential solution to this could be to create a reward mechanism in the crowdsourcing setting, where the reward is determined by both the observed search effectiveness of each human solver, and the uncertainty in the estimation of their search strategy. In the context of BO, this uncertainty can be measured by the covariance of the estimator, i.e., the Hessian of the cost function in Eq.~\eqref{eq:obj}. For people with effective search yet high estimation uncertainty, we can solicit more solutions from them by offering rewards. It would also be interesting to understand the influence of the properties of the problem, e.g., the size of the solution space, on the convergence of the estimation.}

\change{{\bf Knowledge transferability}: The third limitation concerns the transferability of knowledge (search strategies) learned from one task (an optimization problem) to others. This limitation also leads to the question of how ``effectiveness'' of searches shall be measured, as we are not yet able to tell in what condition a strategy that has high rate of improvement (such as P2 in ecoRacer) will continue to produce better solutions than other strategies in a long term.
The same issue, however, exists in IRL: e.g., a control policy learned for pancake flipping does not guarantee optimal egg flipping due to the differences in physical properties between pancakes and eggs. One solution to this in IRL is to allow the policy to adjust to new problem settings, by correcting the state transition model according to the new observations. This solution may also be applied to IBO. In the context of ecoRacer, knowledge such as ``starting acceleration at the beginning of the track'' could be considered as a universal strategy and requires less exploration, while the actual duration for executing this strategy may differ across problem settings. Therefore, it could be more effective for BO to adjust its parameters based on the ones that are learned from human demonstrations on a similar problem, rather than learning from scratch.}

To summarize, IBO could be a valuable tool for machines to mimic human search behavior when (1) the underlying human search mechanism follows BO; (2) the demonstration is sufficient for estimating the true BO parameters with low variances, and (3) the true optimal BO parameters for a long-term search can be estimated based on an effective short-term search.

\cutsubsectionup
\subsection{The difference between learning to search and learning a solution}
\label{sec:mdp}
The proposed IBO approach can be considered as a way to design optimization algorithms with human guidance, and is mathematically similar to IRL. In order to explain the similarities and differences between the two, we first introduce Markov Decision Process (MDP) and Reinforcement Learning (RL), and make an analogy between MDP and an optimization algorithm. 

\cutsubsubsectionup
\subsubsection{Preliminaries on MDP and RL}
A MDP is defined by a tuple $<\mathcal{S}, \mathcal{A}, \mathcal{T}, \mathcal{R}, \gamma, b_0>$ where: $\mathcal{S}$ is a set of states; $\mathcal{A}$ is a set of actions; the state transition function $T({\bf s},{\bf a},{\bf s}')$ determines the probability of changing from state ${\bf s}$ to ${\bf s}'$ when action ${\bf a}$ is taken; $R({\bf s},{\bf a})$ is the instantaneous reward of taking action ${\bf a}$ at state ${\bf s}$; $\gamma \in [0,1)$ is the discount factor of future reward; $b_0({\bf s})$ specifies the probability of starting the process at state ${\bf s}$. In RL, a control policy $\pi$ is a mapping from a state to an action, i.e., $\pi: \mathcal{S} \rightarrow \mathcal{A}$. The long-term {\it value} of $\pi$ for a starting state ${\bf s}$ can be calculated by $V^{\pi}({\bf s}) = R({\bf s},\pi({\bf s})) + \gamma \sum_{{\bf s}'\in {\mathcal S}} T({\bf s},\pi({\bf s}),{\bf s}')V^{\pi}({\bf s}')$, and thus the value of $\pi$ over all possible starting states is the expectation
$V^{\pi} = \sum_{{\bf s}\in \mathcal{S}}b_0({\bf s})V^{\pi}({\bf s})$. A common way to represent a control policy is to introduce a Q-function $Q({\bf s},{\bf a};\boldsymbol{\lambda})$ with unknown control parameters $\boldsymbol{\lambda}$, and let the policy be ${\bf a}({\bf s}) = \text{argmax}_{\mathcal{A}} Q({\bf s},{\bf a};\boldsymbol{\lambda})$. RL identifies the optimal $\boldsymbol{\lambda}$ that maximizes $V^{\pi}$.

\cutsubsubsectionup
\subsubsection{MDP vs. optimization algorithm}
\label{sec:mdpvsop}
An optimization algorithm defines a decision process: Its instantaneous reward is the improvement in the objective value achieved by each new sample, and the cumulative reward represents the total improvement in the objective within a finite number of iterations; its state contains the current solution (in $\mathcal{X}$), the corresponding objective value, and potentially the gradient and higher-order derivatives of the objective function at the current solution; its action is the next solution to evaluate; and its state transition is governed by the optimization algorithm and its parameters. This is similar to MDP where the state transition is affected by the control parameters. The decision process defined by an optimization algorithm, however, is usually non-Markovian, as the new solutions rely on the entire search trajectory. 
Note that it is still possible to consider the optimization process as an MDP, by redefining the state as the continuously growing search trajectory, i.e., elements in the state set $\mathcal{S}$ shall represent all possible search trajectories, rather than samples in $\mathcal{X}$.

\cutsubsubsectionup
\subsubsection{IRL vs. IBO}
RL algorithms identify an optimal control policy for an MDP with a given reward function. However,  real-world applications hardly have explicit definitions of rewards, e.g., the reward for ``driving a car'' cannot be explicitly defined, although people form control policies based on their inherent reward (preference). Therefore, control policy for such applications can be learned more effectively through demonstrations of human beings, which are assumed to be optimal according to the inherent reward of the demonstrator. IRL techniques have thus been developed to identify the reward (and consequently the Q-function and the optimal control policy) that explains human demonstrations, either by estimating the reward parameters so that the demonstrated policy has a higher value than any other policies by a margin~\cite{abbeel2004apprenticeship,ng2000algorithms,ratliff2006maximum,syed2007game}, or by finding the maximum likelihood control parameters directly~\cite{ziebart2008maximum,ramachandran2007bayesian}. 

The IBO approach introduced in this paper is closely related to latter type of IRLs, and more precisely, to the maximum entropy method of Ziebart et al.~\cite{ziebart2008maximum}. Briefly, the maximum entropy IRL proposes the following MLE of parameters $\boldsymbol{\lambda}$ based on a set of demonstrations $h$: 
\cutequationup
\begin{equation}
\begin{aligned}
    \boldsymbol{\hat{\lambda}} &= \text{argmax}_{\boldsymbol{\lambda}} \log P(h|\boldsymbol{\lambda})\\
    &= \text{argmax}_{\boldsymbol{\lambda}} \log\frac{\exp\left(\sum_{({\bf s}_i,{\bf a}_i)\in h} R({\bf s}_i,{\bf a}_i,\boldsymbol{\lambda})\right)}{\prod_{({\bf s}_i,{\bf a}_i)\in h}Z_i(\boldsymbol{\lambda})},
\end{aligned}
    \label{eq:entropy}
\cutequationdown    
\end{equation}
where $Z_i(\boldsymbol{\lambda})$ is a partition function for the visited state ${\bf s}_i$. One can notice the similarities between Eq.~\eqref{eq:entropy} and Eq.~\eqref{eq:obj}: (1) Both are maximum likelihood parameter estimations related to an instantaneous cost, i.e., the reward in Eq.~\eqref{eq:entropy} and the expected improvement in IBO. (2) Both involves partition functions that are computationally expensive, and dependent on the parameters $\boldsymbol{\lambda}$. Due to this dependency, a direct Markov-Chain Monte Carlo (MCMC) sampling in the space of $\boldsymbol{\lambda}$ (e.g., as in \cite{ramachandran2007bayesian}) cannot be applied to optimize the likelihood function since the partition values for two different samples of $\boldsymbol{\lambda}$ do not cancel. Ziebart et al. discussed on alternative approach to address this computational challenge, by using the ``Expected Edge Frequency Calculation'' algorithm that has a complexity of $O(N|\mathcal{S}||\mathcal{A}|)$ for each gradient calculation of the objective in Eq.~\eqref{eq:entropy}, where $N$ is a large number~\cite{ziebart2008maximum}. However, this approach can be infeasible for the IBO estimation problem in Eq.~\eqref{eq:obj} since (1) the space $\mathcal{X}$ is usually continuous, and (2) even with a discretization of $\mathcal{X}$, the enormous size of $\mathcal{S}$ and $\mathcal{A}$ can easily make the calculation intractable, based on the discussion in Sec.~\ref{sec:mdpvsop}. 

\change{Further, one shall notice that IRL and IBO uses different assumptions about human demonstrations: Demonstrations in IRL are assumed to be {\it near-optimal}. Thus learning from them leads to an optimal control policy for an MDP. Demonstrations in IBO, on the other hand, are assumed to be from an effect search strategy, yet are not necessarily optimal. Thus learning from them leads to an optimization algorithm, rather than a solution.} This difference affects the application of the two: IRL can be used when the machine is told to mimic existing solutions, by understanding why these solutions are considered good, e.g., it answers the question ``why do people flip pancakes this way?''; IBO can be used when the machine is meant to mimic the process of searching for good solutions, by understanding how to evaluate the expected improvement of solutions, e.g., it answers the question ``how did people figure out this way of pancake flipping?''.

\cutsectionup
\section{Conclusions}
In this paper, we attempted to address a dilemma in design crowdsourcing: While human beings acquire more advanced intelligence than machines in solving certain types of optimal design problems, soliciting valuable solutions through existing crowdsourcing mechanisms is not cost-effective due to the lack of control over crowd participation and the problem-specific qualification of the crowd. Based on the previous finding that more people acquire good searching strategies than good solutions, we proposed in this paper to \highlight{mimic human search demonstrations} by inversely learn a Bayesian optimization algorithm, so that long-term search can be executed more effectively by the computer even when human solvers abandon the problem. Through simulation and case studies, we showed improved performance of BO when it is equipped with parameters learned through an effective human search. However, the significant performance gap between a human demonstrator and the proposed algorithm in the case study suggested room for improvement of the algorithm. Future investigation will focus on closing this gap by exploring more suitable cognitive models of human solution searching for specific types of optimal design problems.

\cutsectionup
\section*{Acknowledgement}
\label{sec:ack} This work has been supported by the National Science Foundation
under Grant No. CMMI-1266184. This support is gratefully acknowledged.
\bibliographystyle{asmems4}
\bibliography{reference.bib}

\cutsectionup
\section*{Appendix}
\label{sec:appendix}
\subsection*{Derivation of Eq. \eqref{eq:importance}}
Let $p(x)=1/D$ and $q(x)$ be a uniform and a normal density function, respectively, $D$ be the size of $\mathcal{X}$, and $f(x)$ be the function to be integrated. Also let $\mathcal{I}$ and $\mathcal{J}$ be the sample sets drawn from these two distributions, with sizes $I :=|\mathcal{I}|$ and $J :=|\mathcal{J}|$. We have
\cutequationup
\begin{equation*}
\begin{aligned}
\int f(x)dx & = D \int f(x)p(x)dx \\
& = D \left(\int \frac{f(x)p(x)^2}{p(x)+q(x)}dx + \int \frac{f(x)q(x)p(x)}{p(x)+q(x)}dx\right)\\
& \approx D \left(\frac{1}{I} \sum_{\mathcal{I}}\frac{f(x)p(x)}{p(x)+q(x)} + \frac{1}{J} \sum_{\mathcal{J}}\frac{f(x)p(x)}{p(x)+q(x)} \right)\\
& = \sum_{\mathcal{I}}\frac{f(x)D}{I(1+Dq(x))} +  \sum_{\mathcal{J}}\frac{f(x)D}{J(1+Dq(x))}
\end{aligned}
\cutequationdown
\end{equation*}

\cutsubsectionup
\change{\subsection*{IBO behavior under near-random search}
\cutparagraphup
\paragraph{Properties of $l$ and $\tilde{l}$} 
From Sec.~\ref{sec:par}, the unbiased estimation of $l({\bf x}, \alpha_{BO})$ through importance sampling is: 
\cutequationup
\begin{equation}
    \hat{l}({\bf x}, \alpha_{BO}) = -\log \frac{\exp(\alpha_{BO}Q_{EI}({\bf x}))}{\hat{Z}_{BO}/D}.
    \label{eq:conj}
\cutequationdown
\end{equation}
$\hat{l}({\bf x}, \alpha_{BO})$ has the following properties. {\bf Property 1}: $\alpha_{BO}=0$ leads to $\hat{l}({\bf x}, 0)=0$, indicating that ${\bf x}$ is uniformly sampled. One can see that the optimal cost of $L_{BO}$ is non-positive, as one can always achieve $L_{BO}=0$ by considering samples to be uniformly drawn. {\bf Property 2}: When the expected improvement function is constant almost everywhere, i.e., $\text{Pr}(Q_{EI}({\bf x})=C)=1$, we have $\text{Pr}(\hat{l}({\bf x}, \alpha_{BO})=0)=1$. This is because a uniformly drawn initial guess will almost surely satisfy the optimality condition for maximizing a constant function. {\bf Property 3}: Notice that $1+Dq({\bf x}_i)\approx 1$ for ${\bf x}_i\in\mathcal{U}$ due to the small $\sigma_{I}$ (see Sec.~\ref{sec:par}), and $\frac{\exp(\alpha_{BO}Q_{EI}({\bf x}_i))}{1+Dq({\bf x}_i)} \approx 0$ for large $D$ and small $\alpha_{BO}$. The partial derivative of $\hat{l}({\bf x}, 0)$ with respect to $\alpha_{BO}$ can be approximated as:
\cutequationup
\begin{equation}
    \frac{\partial\hat{l}({\bf x}, 0)}{\partial\alpha_{BO}} = c(\alpha_{BO})\sum_{\mathcal{U}}\Delta a_i,
    \label{eq:conjg}
\cutequationdown
\end{equation}
where $c(\alpha_{BO})>0$ and $\Delta a_i := Q_{EI}({\bf x}_i)-Q_{EI}({\bf x})$. Here we need to introduce a conjecture: Let $\bar{Q}_{EI} := \int_{\mathcal{X}} Q_{EI}({\bf x})d{\bf x}/D$ be the average expected improvement, and $A := \int_{\mathcal{X}} \mathbb{1}(Q_{EI}({\bf x})>\bar{Q}_{EI})d{\bf x}$ be the measure of a subspace where the sampled expected improvement value is higher than $\bar{Q}_{EI}$. $A$ decreases from above to below $D/2$ along the increase of the BO sample size. In other words, a uniformly drawn sample has more than $50\%$ of chance to have an expected improvement value higher than $\bar{Q}_{EI}$ at the early stage of BO, and less than $50\%$ at the late stage.}

\change{One evidence of the conjecture is illustrated in Fig.~\ref{fig:bo}: In the first iteration, $\bar{Q}_{EI}$ is slightly lower than $0.5$ while the majority of $\mathcal{X}$ has $Q_{EI}>\bar{Q}_{EI}$; in the fourth iteration, however, only a small region around the peak has $Q_{EI}>\bar{Q}_{EI}$. Using this conjecture, we can show that $\sum_{\mathcal{U}}\Delta a_i<0$ when the sample size is small, thus $\frac{\partial \hat{l}({\bf x}, 0)}{\partial \alpha_{BO}}<0$. Together with Property 1, we have $\hat{l}({\bf x}, \alpha_{BO})<0$ for a small $\alpha_{BO}$ and a small sample size.}

\change{{\bf Property 4}: We notice that in this experiment, the discrepancy between LHS and the modeled max-min sampling scheme leads to overall high (positive) $\tilde{l}$ values, indicating that the samples are not likely follow this scheme. This is consistent with the fact that LHS is not exactly the same as max-min sampling, at least until all of $h_0$ has been considered. We also see that negative $\tilde{l}$ values can be observed when $\alpha_{INI}$ is low, suggesting that the LHS samples can be better explained by a loosely executed max-min sampling scheme than a strict one.}

\cutparagraphup
\change{\paragraph{Discussion on findings from Fig.~\ref{fig:simulation1}}We now summarize a complete list of findings based on these properties. {\bf Finding 1}: A comparison between $\alpha_{INI}=1.0$ and $10.0$ leads to a finding consistent with Property 4. Since the samples are not likely to be drawn from a strictly executed max-min sampling scheme, the entire search trajectory is considered to be created from BO in the case of $\alpha_{INI}=10.0$. While the early samples (less than 10) can be considered as from max-min sampling when $\alpha_{INI}=1.0$ ($\tilde{l}<0$), the low magnitude of $\tilde{l}$ causes this difference to be only visible in the case of $\boldsymbol{\Lambda} =10.0{\bf I}$, where the magnitude of $l$ is also low.
{\bf Finding 2}: IBO correctly identifies the true $\boldsymbol{\Lambda}$ within a few iterations after the initial exploration, except for the case of $\boldsymbol{\Lambda} = 10.0 {\bf I}$.
To explain this exception, we first note that $\boldsymbol{\Lambda}=10.0{\bf I}$
leads to an expected improvement function that is constant almost everywhere (except for the sampled locations where $Q_{EI}=0$) and thus BO reduces to uniform sampling. From Property 2, $L_{BO}=0$ almost surely when we have the correct guess on $\boldsymbol{\Lambda}$. 
Also recall from Property 4 that $L_{INI}>0$ when $\alpha_{INI}$ is high. The above two together explain why with the correct guess of $\boldsymbol{\Lambda} = 10.0 {\bf I}$, we have $L$ close to zero when $\alpha_{INI}=10.0$ and slightly negative when $\alpha_{INI}=1.0$.\footnote{But why does the guess of $\boldsymbol{\Lambda} = 10.0 {\bf I}$ lead to significantly decreasing $L$ in the other three cases? This is because in those cases, BO does not resemble random sampling, i.e., the sequences of samples are more clustered. When a new sample is among this cluster, its similarities to existing ones are non-zero even when a large $\boldsymbol{\Lambda}$ is assumed, due to the small Euclidean distance among the pairs. And in turn, the expected improvement function has peaks within the clusters, and remains constant far away from them, rather than being a constant almost everywhere. As a result, the optimal value of $\hat{l}({\bf x}, \alpha_{BO})$ with respect to $\alpha_{BO}$ becomes negative, even when $\boldsymbol{\Lambda}$ is incorrectly guessed as $10.0 {\bf I}$.}

To explain the negative $L$ values for the incorrect guesses of $\boldsymbol{\Lambda}$, we use Property 3 to show that when the sample size is small and the expected improvement function is not flat, $L_{BO}<0$ for a small $\alpha_{BO}$, and thus $L<0$. To summarize, Finding 2 suggests that for a search trajectory with a limited length that resembles a random search, the proposed IBO approach will consider it being derived from a BO that loosely solves Eq.~\ref{eq:obj}. However, this caveat is of little practical concern, since (1) a random search rarely outperforms BO with non-trivial settings, and (2) a BO with low $\alpha_{BO}$ (instead of high $\boldsymbol{\Lambda}$) can equally simulate a random search.} 

\end{document}